\documentclass[10pt,twocolumn,letterpaper]{article}

\usepackage{cvpr}
\usepackage{times}
\usepackage{epsfig}
\usepackage{graphicx}
\usepackage{subcaption}
\usepackage{algorithm}
\usepackage{algorithmic}
\usepackage{setspace}
\usepackage{multirow}
\usepackage{amsmath}
\usepackage{amssymb}

\usepackage[pagebackref=true,breaklinks=true,letterpaper=true,colorlinks,bookmarks=false]{hyperref}
\usepackage{cleveref}

\cvprfinalcopy 

\def\cvprPaperID{2} 

\ifcvprfinal\fi
\begin{document}

\title{Real-time Pupil Tracking from Monocular Video for Digital Puppetry}

\author{Artsiom Ablavatski \quad Andrey Vakunov \quad Ivan Grishchenko \quad Karthik Raveendran \quad Matsvei Zhdanovich\\
Google Research\\
1600 Amphitheatre Pkwy, Mountain View, CA 94043, USA\\
{\tt\small \{artsiom, vakunov, igrishchenko, krav, matvey\}@google.com}}

\maketitle
\begin{abstract}
We present a simple, real-time approach for pupil tracking from live video on mobile devices. Our method extends a state-of-the-art face mesh detector with two new components: a tiny neural network that predicts positions of the pupils in 2D, and a displacement-based estimation of the pupil blend shape coefficients. Our technique can be used to accurately control the pupil movements of a virtual puppet, and lends liveliness and energy to it. The proposed approach runs at over $50$ FPS on modern phones, and enables its usage in any real-time puppeteering pipeline.
\end{abstract}

\section{Introduction}
The task of animating a virtual puppet in real-time using live footage of a human is a well studied one. Broadly speaking, one can classify these techniques by their choice of input data (monocular video, multi-view, depth images) and the methodology (direct optimization, prediction using neural networks, heuristics). For instance, Ichim~\etal ~\cite{ichim2015dynamic} use monocular videos with predefined camera movements in order to obtain dense registration of person specific facial features and create a dynamic face model on the fly via optimization. Wu~\etal~\cite{wu2019mvf} leverage multi-view data to align a 3D Face Morphable Model (3DMM~\cite{blanz1999morphable}) using a bundle of neural networks and produce person specific blend shapes. We refer the reader
to~\cite{thies2016face2face} for a review of related work on 3D face alignment and blend shape computation. In this paper, we focus on puppeteering on mobile devices, without the use of extra sensors or a person-specific calibration step.

Despite the success of these techniques, the resulting avatars tend to lack a certain liveliness or expressivity because they do not track the position of the pupils. For instance, prior approach~\cite{kartynnik2019real} leverage 3DMM which does not have the pupils in its internal representation. We address this problem using a two stage pipeline that combines a neural network for predicting the position of the pupils (\Cref{sec:pipeline}) and a displacement-based algorithm for estimating the pupil blend shapes (\Cref{sec:pipeline}). We build this pipeline on top of a state-of-the-art face mesh prediction model~\cite{kartynnik2019real}, but our approach generalizes to other face meshes.

Our network detects 5 points of the pupil, outer iris circle, and eye contour for each eye. Based on the position of these points, we apply carefully devised heuristics to obtain blend shape coefficients in the range $[-1, 1]$ where 1 and -1 represent full blend shape activation (e.g. the eye looks up or down respectively) with 0 being the neutral position (the eye looks frontally). We follow this with post-processing to reduce jitter from the detection stage and make the final rendering smooth and appealing.

\begin{figure}[tb]
    \centering
    \begin{subfigure}[b]{0.49\columnwidth}
        \frame{\includegraphics[width=\linewidth]{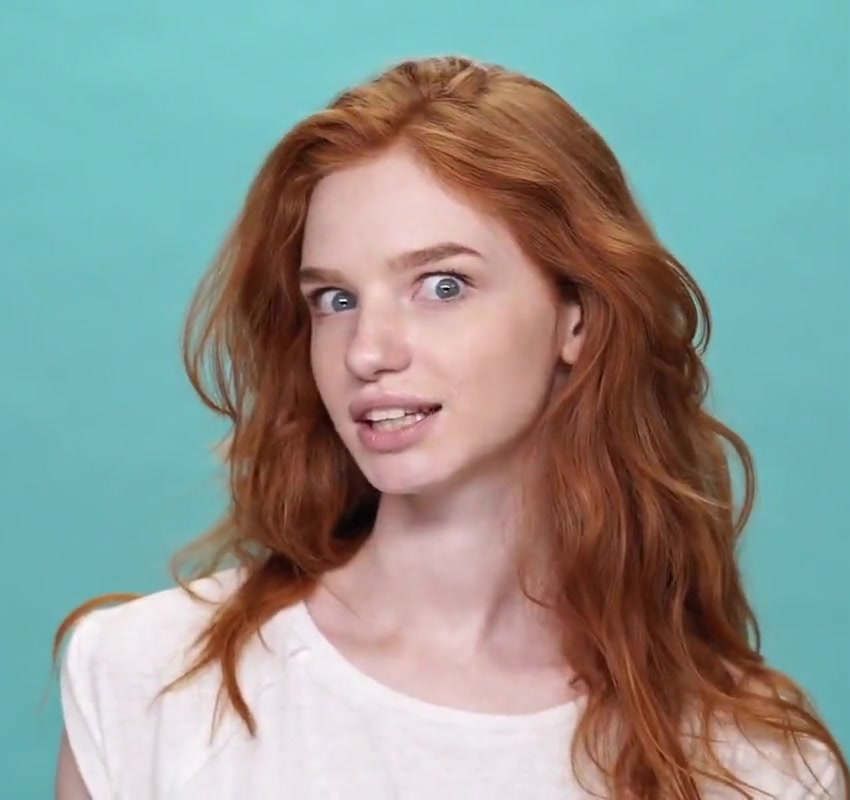}}
    \end{subfigure}
    \begin{subfigure}[b]{0.49\columnwidth}
        \frame{\includegraphics[width=\linewidth]{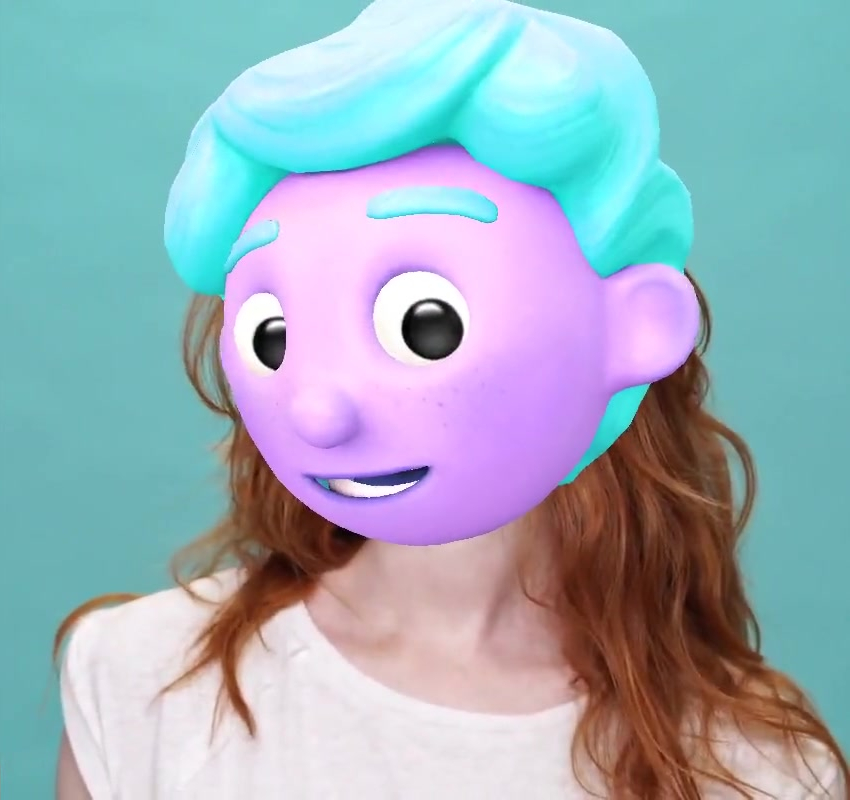}}
    \end{subfigure}
    \caption{Final rendering of eyes blend shapes tracking on the virtual avatar. Left --- the original image, right --- the image with overlaid avatar driven by the acquired blend shapes.}
    \label{fig:examples}
\end{figure}

Our approach only requires a single frame at a time and does not rely on any additional sensors such as a depth camera. \Cref{fig:examples} shows an example of a virtual puppet animated with our technique.

\section{Neural network based eye landmarks} \label{sec:pipeline}

\begin{figure*}[ht]
    \centering
    \includegraphics[width=\linewidth]{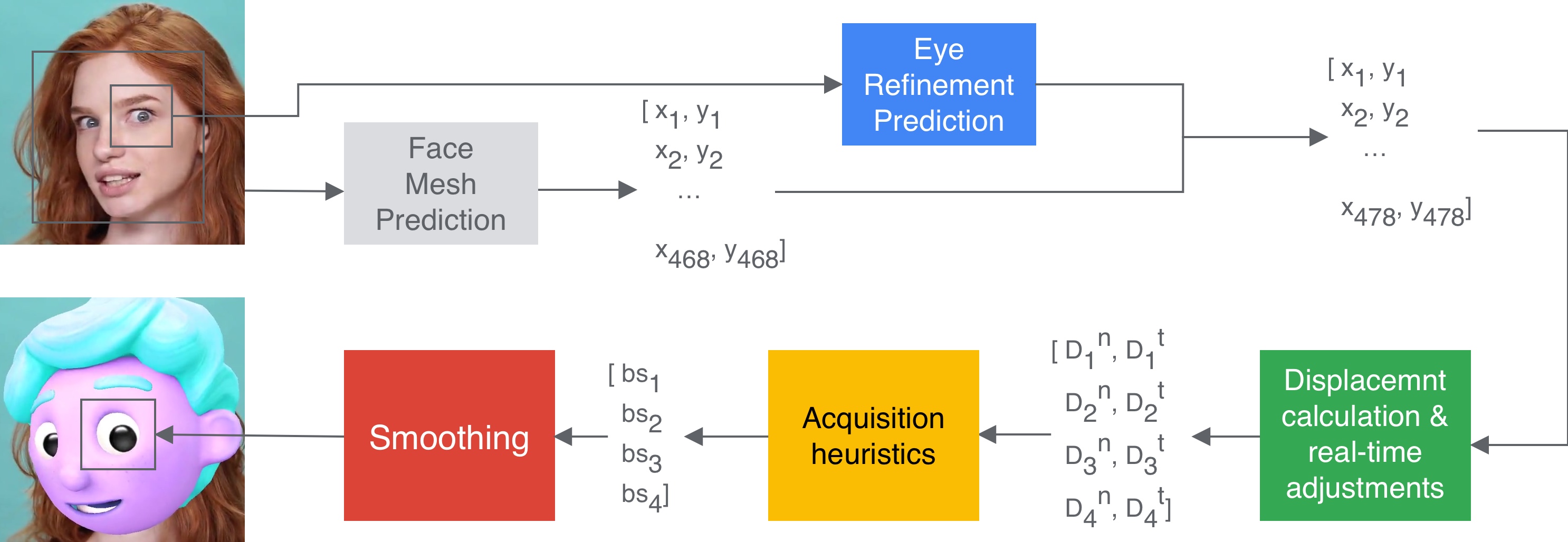}
    \caption{Overview of the pupil blend shapes acquisition. See text for details.}
    \label{fig:overview}
\end{figure*}

We start with a modern face mesh estimation pipeline that predicts a 468 vertex mesh for the human face \cite{kartynnik2019real}. We then compute the bounding boxes for eye regions and pass the corresponding cropped image regions to a smaller landmark regression network that produces additional higher quality landmarks. 

Specifically, we extract the corresponding region ($64\times64$ pixels) via cropping from the center of eye landmark of the face mesh estimator. This cropped region is fed into a tiny neural network that has a structure similar to that described in Bazarevsky~\etal~\cite{bazarevsky2019blazeface}. This subsequent network predicts 5 locations in 2D (pupil center, 4 points of outer iris circle, and 16 points of eye contour) in the coordinate system of the image starting from the upper left corner. We combine the corresponding landmarks (16 points of eye contour) from the face estimation pipeline with those from the eye refinement network by replacing the $x, y$ coordinates of the former while leaving $z$ untouched. We extend the face mesh with 5 pupil landmarks (pupil center and 4 points of outer iris circle), with their $z$ coordinate set to the average of the $z$ coordinate of the eye corners. The final refined facial mesh contains 478 (468 landmarks + 5 left pupil landmarks + 5 right pupil landmarks) vertices and is used in the second stage of this pipeline.

To execute the neural network on mobile devices, we employ TensorFlow Lite with GPU backend~\cite{lee2019device} coupled with MediaPipe~\cite{lugaresi2019mediapipe} --- a framework for building perception pipelines.

\subsection{Model architecture}

The neural network for predicting eye and iris landmarks contains a number of bottlenecks similar to recent work of Tan~\etal~\cite{tan2019mnasnet} and Bazarevsky~\etal~\cite{bazarevsky2019blazeface}. 
The model ends with a fully connected layer that outputs a $5\times2$ tensor corresponding to $x,y$ coordinates for each landmark (defined in the cropped image coordinate system). This design allows the network to learn a rich feature representation that achieves low error rates per landmarks (see~\Cref{sec:results}). Further, the proposed model architecture has a small memory foot print (due to the small input resolution) and low number of FLOPs (due to the $1\times1$ compression and $3\times3$ depthwise convolutions). This enables real-time performance of the network on CPU and super real-time performance using GPU capabilities on modern phones. The run-time measurements and error rates are shown in the~\Cref{tab:quantitative_results}.

\section{Displacement-based pupil blend shape estimation}
We use the refined mesh to predict 4 blend shapes for the pupils: pupils pointing outwards, inwards, upwards and downwards respectively. We compute the activation of these shapes using a simple yet powerful displacement based approach. Specifically, for every blend shape we choose a pair of vertices on the refined mesh that robustly captures the blend shape i.e. for the pupil pointing inwards, we use the vertex of the pupil and the vertex of eye corner. Next, we measure the displacement $D_{\mathit{current}}$ between these two vertices and compare it to two empirically derived displacements $D_{\mathit{neutral}}$ --- the displacement with the minimum activation of the blend shape, and $D_{\mathit{activated}}$ --- the displacement measured using the maximum activation of the blend shape. Based on this comparison, we obtain a scalar value in the range of $[0, 1]$ for each pupil blend shape.

Next, we merge pairs of opposite blend shapes into two aggregate blend shapes. Finally, we apply smoothing and couple the estimated blend shape values between both eyes.

An overview of the pipeline is presented on the \Cref{fig:overview}.

\subsection{Real-time heuristics calibration}

\begin{figure}[h]
    \centering
    \includegraphics[width=\linewidth]{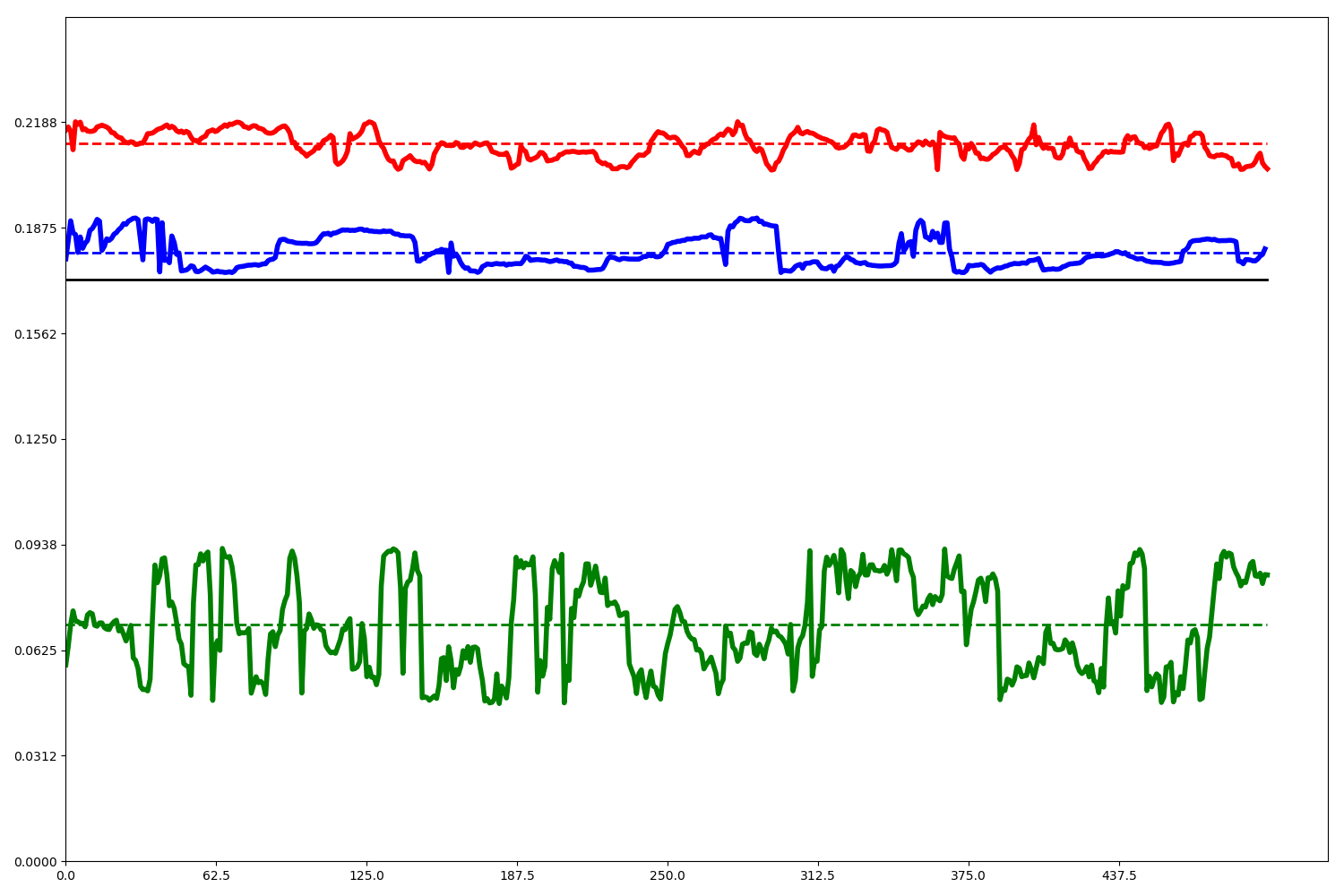}
    \caption{Variation of $D_{\mathit{neutral}}$ displacements in time for 3 subjects (red, green, blue) and initial estimated value (black). Vertical axis --- $D_{\mathit{neutral}}$ displacement values from 25th to 75th percentile. Horizontal axis --- \# frames}
    \label{fig:displacements}
\end{figure}

Heuristics play a vital part in the proposed blend shape pipeline. Algorithm described in the~\Cref{sec:pipeline} requires two displacements: $D_{\mathit{neutral}}$ and $D_{\mathit{activated}}$ to be defined. The initial displacements are empirically estimated based on the representative face mesh dataset. However, these initial values are unable to model all person-specific variations. For a visual reference,~\Cref{fig:displacements} shows the variation in displacements $D_{\mathit{neutral}}$ over time for different subjects (drawn in different colors) and initial estimated value (drawn in black). The solid line represents measurements of the actual displacements on per frame basis while the dotted line indicates the smoothed trend for a specific subject.

To address the challenge of person-specific displacements and to make the system reliable, we propose to enhance the displacement estimation with a real-time calibration step. We employ the standard score~\cite{yin2017case} calculation algorithm with a few modifications. The main idea of the filter is to check the displacement on every iteration and add it to a circular buffer of the trusted displacements if it falls within the specified confidence interval. Consequently, the calibrated displacement is calculated as an average of the trusted displacements. The standard deviation of these trusted displacements is used as the confidence interval in the next iteration. Details of the Standard Score Filter algorithm are presented in~\Cref{alg:standard_score}.

\begin{algorithm}
        \caption{Standard Score Filter}
        \label{alg:standard_score}
        \begin{algorithmic}
            \REQUIRE $D_{\mathit{initial}},D_{\mathit{current}},Thrs_{\mathit{variance}}$
            \REQUIRE $F_{\mathit{influence}},F_{\mathit{annealing}}$
            \STATE $Diff \Leftarrow D_{\mathit{current}} - D_{\mathit{mean}}$
            \STATE $Interval \Leftarrow Thrs_{\mathit{variance}} * Variance_D$
            \IF{$Diff \in Interval$}
                \STATE $D_{\mathit{trusted}} \Leftarrow D_{\mathit{current}}$
            \ELSE
                \STATE $\alpha \Leftarrow F_{\mathit{influence}}$
                \STATE $\beta \Leftarrow 1-F_{\mathit{influence}}$
                \STATE $D_{\mathit{trusted}} \Leftarrow \alpha * D_{\mathit{current}} + \beta * D_{\mathit{mean}}$
            \ENDIF
            \STATE  $D_{\mathit{trusted\_list}}.insert(D_{\mathit{trusted}})$
            \STATE $F_{\mathit{influence}} \Leftarrow max(0, F_{\mathit{influence}} - F_{\mathit{annealing}})$
            \STATE $D_{\mathit{calibrated}} \Leftarrow mean(D_{\mathit{trusted\_list}})$
            \STATE $Variance_D \Leftarrow std(D_{\mathit{trusted\_list}})$
            \RETURN $D_{\mathit{calibrated}}$
        \end{algorithmic}
    \end{algorithm}

\section{Datasets and training}
In order to train the neural network to infer 2D positions of points around the eye, we use $\approx 20,000$ manually annotated images from a globally sourced dataset. We applied a set of augmentations to these images such as affine (rotation, flip) and color transformations (hue, saturation, non-linear mapping, realistic camera noise injection). The network was trained for 250 epochs using the Adam optimizer~\cite{kingma2014adam}. Similar to~\cite{kartynnik2019real}, we use the Mean Squared Distance normalized by the Inter-Eye Distance (MSE IED) as our loss function. This normalization avoids factoring in the scale of the eyes. 

\section{Results} \label{sec:results}

To quantitatively estimate the accuracy of the trained model, we use  $\approx 4000$ manually annotated images. The trained model achieved 7.16\% MAD IED (Mean Absolute Distance normalized by the Inter-Eye Distance) on the collected dataset. The baseline error of the manual annotation is 5.73\% for simple use cases (the face on the image is frontally rotated) and 7.04\% - for hard cases. The error was measured on the same images annotated by different subjects. The inference speeds of the face mesh as well as eye refinement models on a number of phones are shown in the~\Cref{tab:quantitative_results}.

\begin{table}[h]
\centering
\begin{tabular}{|l|c|c|c|}
\hline
\multirow{3}{*}{Phone} & \multicolumn{3}{c|}{Inference speed (ms)}  \\ \cline{2-4}
& Face mesh &  \multicolumn{2}{c|}{Eye refinement}  \\ \cline{2-4}
& GPU & ~~~~CPU~~~~ & ~~~~GPU~~~~ \\ \hline
Pixel XL & 14 & 16 & 12 \\ \hline
Pixel 2 XL & 12 & 20 & 8 \\ \hline
Pixel 3 XL & 10 & 12 &  5  \\ \hline
Samsung S9 & 10 & 12 & 5   \\ \hline
iPhone X & 4 & 7 & 2.6  \\ \hline
\end{tabular}
\vskip 1ex
\caption{Face mesh and Eye refinement models inference speeds on a number of phones.}
\label{tab:quantitative_results}
\end{table}
\section{Conclusion}

We present a novel pipeline for real-time pupil tracking from live video on mobile devices at real-time speeds. The approach defines a full end-to-end pipeline for pupil blend shapes estimation from monocular images without any pre-calibration and can be combined with any existing blend shape implementation. It can be used as out-of-the box solution for accurate control of the eye movements for a virtual puppet.

{\small
\bibliographystyle{ieee_fullname}
\bibliography{paper}
}

\end{document}